\def\year{2022}\relax
\documentclass[letterpaper]{article} 
\usepackage{aaai22}  
\usepackage{times}  
\usepackage{helvet}  
\usepackage{courier}  
\usepackage[hyphens]{url}  
\usepackage{graphicx} 
\def\UrlFont{\rm}  
\usepackage{natbib}  
\usepackage{caption} 
\usepackage{subcaption}
\usepackage{multirow}
\usepackage{tabularx}
\usepackage{array}
\usepackage{makecell}
\usepackage{caption} 
\usepackage{tcolorbox}
\usepackage{amsmath}
\DeclareCaptionStyle{ruled}{labelfont=normalfont,labelsep=colon,strut=off} 
\usepackage{algorithm}
\usepackage{algorithmic}

\usepackage{newfloat}
\usepackage{listings}

\title{Heterogeneous Graph Neural Architecture Search with GPT-4}

\author{Haoyuan Dong\textsuperscript{\rm 1}\equalcontrib, 
Yang Gao\textsuperscript{\rm 2}\equalcontrib, 
Haishuai Wang\textsuperscript{\rm 2},
Hong Yang\textsuperscript{\rm 1}, 
Peng Zhang\textsuperscript{\rm 1}\thanks{Corresponding author.}\\}
\affiliations{
\textsuperscript{1}Cyberspace Institute of Advanced Technology, Guangzhou University, China\\
\textsuperscript{2}College of Computer Science and Technology, Zhejiang University, China\\
2112233157@e.gzhu.edu.cn
gaoyang9507@zju.edu.cn \ haishuai.wang@zju.edu.cn \  hyang@gzhu.edu.cn \ p.zhang@gzhu.edu.cn\\
}

\begin{document}

\maketitle

\begin{abstract}
 Heterogeneous graph neural architecture search (HGNAS) represents a powerful tool for automatically designing effective heterogeneous graph neural networks. However, existing HGNAS algorithms suffer from inefficient searches and unstable results. In this paper, we present a new large language model based HGNAS model to improve the search efficiency and search accuracy of HGNAS. Specifically, we present a new GPT-4 enhanced Heterogeneous Graph Neural Architecture Search  (\textbf{GHGNAS} for short). The basic idea of GHGNAS is to design a set of prompts that can guide GPT-4 towards the task of generating new heterogeneous graph neural architectures. By iteratively asking GPT-4 with the prompts, GHGNAS continually validates the accuracy of the generated HGNNs and uses the feedback to further optimize the prompts. Experimental results show that GHGNAS can design new HGNNs by leveraging the powerful generalization capability of GPT-4. Moreover, GHGNAS runs more effectively and stably than previous HGNAS models based on reinforcement learning and differentiable search algorithms.
\end{abstract}

\section{Introduction}
Heterogeneous graphs are often used to describe complicated relationships between different types of objects, such as social networks~\cite{jacob2014learning} and recommendation networks~\cite{dong2012link}. Recently, a new type of Heterogeneous Graph Neural Network models (HGNNs) are proposed to capture rich semantic information in heterogeneous graphs~\cite{schlichtkrull2018modeling}. The main idea of  HGNNs is to iteratively update node representations by aggregating information from either neighboring nodes or high-order nodes generated from meta-paths~\cite{sun2011pathsim}. Typical HGNNs include 
 RGCN~\cite{schlichtkrull2018modeling}, HAN~\cite{wang2019heterogeneous}, and MAGNN~\cite{fu2020magnn}. 

Recently,  Neural Architecture Search (NAS) algorithms ~\cite{zoph2016neural,elsken2019neural} are introduced into HGNNs which can automatically design neural architectures for HGNNs without human manual intervention. For example, an early work HGNAS~\cite{gao2021heterogeneous} uses reinforcement learning to find the best neural architectures for HGNNs. Based on HGNAS, a number of new neural architecture search methods are proposed for heterogeneous graphs. For example, AutoGEL~\cite{wang2021autogel}, MR-GNAS~\cite{zheng2022multi} and DHGNA~\cite{zhang2023dynamic} optimize a supernet with differentiable search algorithms~\cite{liu2018darts,xie2018snas}, and Space4HGNN~\cite{zhao2022space4hgnn} introduces random search based on a modular search space. 
Moreover, a number of models focus on designing meta-structures~\cite{huang2016meta} which can be taken as generalized forms of meta-paths to describe relationship patterns between nodes in heterogeneous graphs. For instance, GEMS~\cite{han2020genetic} and DiffMG~\cite{ding2021diffmg} use evolutionary algorithms~\cite{vcrepinvsek2013exploration} and differentiable algorithms respectively to find the best meta-structures for downstream graph learning tasks.

While current HGNAS methods can discover more superior HGNN models than manual design, they still necessitate a considerable amount of expertise to construct the search space and optimize search strategies. Notably, the aforementioned methods are influenced to some extent by random seeds~\cite{kazimipour2014review,henderson2018deep}. For instance, the differentiable search method DiffMG~\cite{ding2021diffmg} typically demonstrates unstable search results because the performance of differentiable search algorithms depends greatly on the selection of the search seed. Choosing inappropriate seeds may lead to search results inferior to manually designed HGNN models. Therefore, while these methods have provided effective approaches for designing heterogeneous graph neural networks, reducing reliance on expertise and mitigating the impact of random seeds remains an open issue that needs to be addressed before they can be more robustly and widely applied.

Large Language Models (LLMs)~\cite{zhao2023survey} such as GPT-4 have shown powerful generation capability and have been used in neural architecture search in recent years. For example, AutoML-GPT~\cite{zhang2023automl} uses GPT to handle data processing, hyper-parameter tuning, and model training. GENIUS~\cite{zheng2023can} uses GPT-4 to find CNN architectures for downstream tasks. GPT4GNAS~\cite{wang2023graph} uses GPT-4 to generate new homogeneous graph architectures. However, these methods have not considered heterogeneous graphs which require a more complicated search space. In order to enable GPT-4 for heterogeneous graphs, \textit{the key issue is to design a new class of prompts that can guide GPT-4 to iteratively generate better HGNN architectures}.

In this paper, we present a new Heterogeneous Graph Neural Architecture Search method based on the large language model GPT-4, namely \textbf{GHGNAS}. Our method designs a new class of prompts for GPT-4 which can help generate new architectures for HGNNs that are suitable for downstream graph learning tasks. Specifically, the prompts consist of descriptions of the learning task and input heterogeneous graphs, and descriptions of the search space, search strategy, and search feedback of the HGNAS method. These prompts can guide GPT-4 to adapt to the task of generating new neural architectures for HGNNs. Then, the evaluation result of each new HGNN architecture is taken as feedback which will be added to the prompt for generating better HGNN architectures. By repeatedly updating the prompts, GHGNAS can obtain the best HGNN architectures. We summarize the contribution of the paper as follows:
\begin{itemize}
    \item This research represents the first work on using GPT-4 to generate new heterogeneous graph neural architectures, where a new heterogeneous graph neural architecture search method GHGNAS is presented. 

    \item This work proposes a new class of prompts to enable GPT-4 for heterogeneous graph neural network architecture search. These prompts can guide GPT-4  to iteratively generate better HGNN architectures for downstream learning tasks.
    \item Experiments on benchmark datasets show that the proposed GHGNAS method can obtain more accurate and stable results than existing methods on heterogeneous graph neural architecture search
\end{itemize}

\section{RELATED WORK}
In this part, we survey related work on heterogeneous graph neural architecture search and large language models for neural architecture search.

\subsection{Heterogeneous graph neural architectures search}

Existing methods for heterogeneous graph neural architecture search can be broadly categorized into three classes, i.e., reinforcement learning algorithms, differentiable search algorithms, and evolutionary search algorithms. 

HGNAS \cite{gao2021heterogeneous} represents the first work of using reinforcement learning to heterogeneous graph neural architecture search. Based on HGNAS, HGNAS++~\cite{gao2023hgnas++} is proposed to accelerate the search process. These two methods use reinforcement learning and RNN to first represent a neural architecture as a sequence and then allow the RNN to act as controller to sample architectures, where the testing results of the sampled architectures are taken as rewards to update the weights of the RNN controller. As a result, the RNN controller iteratively generates better heterogeneous graph neural architectures. 

The second type of method is based on differentiable search algorithms, such as DARTS~\cite{liu2018darts}. First, a super network containing all possible operations is constructed, and then the weights of these operations are updated through gradient descent. The final neural network structure is composed by selecting the operation with the largest weight. To solve the problem of graph data heterogeneity, DHGNA~\cite{zhang2023dynamic} adds relationships and timestamps that need to be focused on into the search space to construct a super network, thereby solving the architecture search problem of dynamic heterogeneous graphs, MR-GNAS~\cite{zheng2022multi} sets up a fine-grained search space for multi-relationship networks, AutoGEL~\cite{wang2021autogel} incorporates additional operators into the search space to handle edge
features. Different from directly searching for architectures, the goal of DiffMG~\cite{ding2021diffmg} and PMMM~\cite{li2023differentiable} is to explore and find better meta-structures.

The third type of method is based on evolutionary algorithms~\cite{vcrepinvsek2013exploration}. This method first initializes a batch of architectures as a population, with each architecture as an individual. This process is similar to the evolution process of organisms, using selection, recombination, and mutation operations to generate a new population, retaining good architectures and eliminating poorly performing ones. A representative work GEMS~\cite{han2020genetic} retains valuable meta-structures by using genetic algorithms.

However, the above works haven't considered using large language models for heterogeneous graph neural architecture search, which is the focus of this paper.

\subsection{GPT-4 for neural architecture search}
Recently, there have been several works proposed to use the GPT-4 for neural architecture search. Thanks to the powerful generation capability of GPT-4, a pioneer work GENIUS~\cite{zheng2023can} uses GPT-4 to design neural architectures for CNNs. The key idea is to allow GPT-4 to learn from the feedback of generated neural architectures and iteratively generate better ones. Experimental results on benchmarks demonstrate that GPT-4 can find top-ranked architectures after several repeats of prompts. Subsequently, AutoML-GPT~\cite{zhang2023automl} designs prompts for GPT to automatically complete tasks of data processing, model architecture design, and hyper-parameter tuning. There are also articles exploring the mutual improvement between LLMs and AutoML~\cite{tornede2023automl}. 

Despite the above achievements, existing works have not explored GPT-4 for heterogeneous graph neural architecture search. The search space for heterogeneous graph neural architecture search is vast which requires considering different ways of message propagation through different graph relationships. This paper designs a new set of prompts for a large language model GPT-4 to enable heterogeneous graph neural architecture search, which represents the first effort at the combination of GPT-4  and heterogeneous graph neural architecture search. 

\begin{figure*}[htb]
    \centering
    \includegraphics[width=0.9\textwidth]{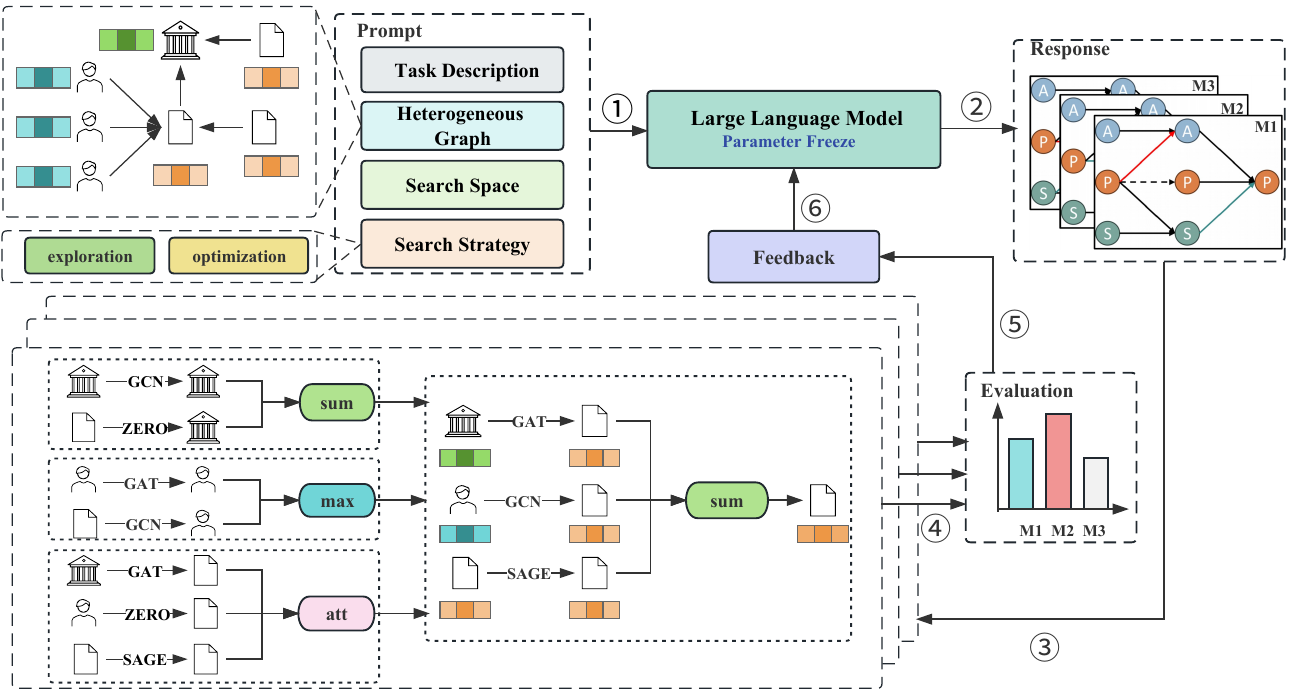}
    \caption{An overview of the proposed GHGNAS method, which includes \textcircled{1} design a prompt containing task description, information about heterogeneous graph, search space, and search strategy; \textcircled{2} based on the prompt, GPT-4 generates a batch of candidate architectures; \textcircled{3} convert the architectures into HGNNs; \textcircled{4} verify the performance of the architectures in a downstream graph learning task; \textcircled{5} generate feedback prompts based on the architectures and accuracy, and \textcircled{6} GPT-4 continues to generate new neural architectures based on the feedback. }
    \label{fig:1}
\end{figure*}

\section{PRELIMINARIES}
\textbf{Heterogeneous graph} $G=\{\mathcal{V},\mathcal{E},\mathcal{T},\mathcal{R},f_{\mathcal{T}},f_{\mathcal{R}}\}$, where $\mathcal{V}$ represents a set of nodes,$ \mathcal{E}$ represents a set of edges, $\mathcal{T}$ denotes a set of node types, and $\mathcal{R}$ denotes a set of edge types. The mapping functions $f_\mathcal{T}(\mathcal{V})\to\mathcal{T}$ and $f_\mathcal{R}(\mathcal{E})\to\mathcal{R}$ map nodes and edges to their respective types. Here, $|\mathcal{T}| + |\mathcal{R}| > 2$.\\
\textbf{Meta-paths} refer to paths from one node to another, which can be represented as $m = \begin{aligned}t_1&\overset{r_1}\to t_2\cdots\overset{r_k}{\to}t_{k+1}\end{aligned}$, where $t.\in\mathcal{T}$ and $r.\in\mathcal{R}$. Meta-paths are used to represent specific relationship patterns.\\
\textbf{Heterogeneous graph neural architecture search.} Given a heterogeneous graph $\mathcal{G}$, the  objective is to find the best architecture  from a given architecture space $\mathcal{A}$ that maximizes a performance evaluation metric $c$ as follows: 
\begin{equation}
\label{eq1}
\alpha^{*} = \underset{\alpha\in\mathcal{A}}{\operatorname*{argmax}}  \ c(\alpha(\mathcal{G})),
\end{equation}
where $\alpha^{*} \in \mathcal{A}$ represents the discovered optimal architecture.  The performance evaluation metric $c$ can be either AUC or macro-F1, depending on the choice of downstream learning tasks.

\begin{table}[b]
\caption{Search space of HGNAS.}
\begin{tabular}{c|ccc}
\hline
Search space & \multicolumn{3}{c}{Operations} \\ \hline
$R^l$& \multicolumn{3}{c}{$R^l \subseteq R \cup  R_{I}, \quad 0 \leq l < n$} \\ \hline
HAggr & \multicolumn{3}{c}{sum, mean, max, lstm, att} \\ \hline
\multirow{6}{*}{GNN} & \multicolumn{1}{c|}{} & \multicolumn{1}{c|}{Encd} & Aggr \\ \cline{2-4} 
 & \multicolumn{1}{c|}{$gcn$} & \multicolumn{1}{c|}{$\frac{X_{s}W_{C}}{\sqrt{|N(s,C)||N(t,C)|}}$} & sum \\ \cline{3-4} 
 & \multicolumn{1}{c|}{$gat$} & \multicolumn{1}{c|}{$X_{s}{W}_{C}$} & att \\ \cline{3-4} 
 & \multicolumn{1}{c|}{$edge$} & \multicolumn{1}{c|}{$(X_{s}-X_{t})W_{C}$} & max \\ \cline{3-4} 
 & \multicolumn{1}{c|}{$sage$} & \multicolumn{1}{c|}{$X_{s}W_{C}$} & max \\ \cline{3-4} 
 & \multicolumn{1}{c|}{$zero$} & \multicolumn{1}{c|}{0} & sum \\ \hline
\end{tabular}
\label{ops}
\end{table}

\textbf{Search space.} We use the default search space of HGNAS, as shown in  Table~\ref{ops}. The search space is designed based on message passing in heterogeneous graphs. Specifically, message passing can be represented as the following parts:
\begin{equation}
M_{r}^{l+1} \leftarrow GNN(X^l,A^r)
\end{equation}
\begin{equation}
X^{l+1}\leftarrow HAggr(\{M_r^{l+1}\mid {\forall}r\in R^l\}), 
\end{equation}
where $X^{l}$ represents the node feature of nodes at the $l-th$ layer, $M_r$ represents the aggregated message vector on relation $r$, $A^r$ represents the adjacency matrix of relation $r$, and $R^l$ represents the set of relationships at layer $l$ that participate in message passing. $GNN$ is used to calculate message vectors from certain types of relationships, and $HAggr$ is employed for aggregating messages from different $r$. 
As shown in Table~\ref{ops},  candidate $GNN$ operations include "$gcn$"\cite{kipf2016semi}, "$gat$"\cite{velickovic2017graph}, "$edge$"\cite{xu2018representation}, "$sage$"\cite{hamilton2017inductive},  "$zero$", and candidate $HAggr$ includes $sum$, $mean$, $max$, $lstm$, $att$, where $att$ includes learnable parameters. 
Note that our approach can also be combined with other search spaces, such as the search space from DiffMG~\cite{ding2021diffmg}. 


\section{Method}
\begin{table}[t]
\caption{The prompts of GHGNAS.}
\begin{tabular}{|m{2cm}|m{5.7cm}|}
\hline
Prompt & Template \\ \hline
\centering Task description $P_t$& Our task is heterogeneous graph neural architecture, searching for a HGNN architecture that can achieve the best performance on a downstream task \\ \hline
\centering Heterogeneous graph $P_d$ & The data set is {[}\textbf{Dataset}{]} ,{[}\textbf{Node type and number}{]}, {[}\textbf{Edge type and number}{]} . The downstream task is {[}\textbf{NC}\textbackslash{}\textbf{LP}{]} and the target \textbf{node\textbackslash{}link} is {[}\textbf{Target}{]} \\ \hline
\centering Search space $P_s$ & The architecture of a n-layer HGNN is expressed as [\textbf{ArchSeq}], for each edge, you need to select one from {[}\textbf{GNN Operations}{]}, then choose an aggregate function from {[}\textbf{HAggr Operations}{]} per layer \\ \hline
\centering Search strategy $P_e \&P_o $  & \makecell[l]{Exploration Strategy:\\Explore as many different architectures in\\ the search space as possible\\Optimization Strategy:\\ Analyze how to get a better architecture\\ based on existing results}\\ \hline
\centering Feedback $P_f$ & The performance of {[}\textbf{Architecture}{]} is {[}\textbf{Accuracy}{]} \\ \hline
\end{tabular}
\label{prompt}
\end{table}

In this section, we introduce the proposed model  GHGNAS in detail. First, we introduce the new prompts. Then, we introduce the search process, where the GPT-4 model is taken as the controller and generates new architectures.

\subsection{GHGNAS Prompts}
In order to solve Eq\ref{eq1}, we need to use natural language to describe the task of neural architecture search. In the following, we discuss two key issues.

\textbf{First, how to let GPT-4 understand the search space and output the architecture?} According to the search space discussed in the previous section, we need to choose an HGNN for each relationship to calculate $M_r$, and choose an aggregation method for all the messages within the layer. An intuitive approach is to represent all relationship types involved in message passing and the aggregation function of each layer as a sequence. Then, the above process is equivalent to asking GPT-4 to choose a specific value for each element on the sequence. For each candidate HGNN, we do not specifically explain their calculation method but emphasize the meaning of "$zero$", to help GPT-4 filter the edges involved in message passing. In the end, we just need GPT-4 to output the sequence representation of the architecture.

\textbf{Second, how to avoid exploring redundant architectures and improve search efficiency?} After giving the search space and the output, GPT-4 already can generate new architectures. However, if GPT-4 is only given a full view of the search space, they will generate in a random, which leads to random search and probably repeat the same architecture in the worst case. Therefore, we need to specify a search strategy to avoid random search.

Our idea is the same as reinforcement learning~\cite{sutton2018reinforcement}, considering two strategies of exploration and exploitation. We divide the search into two stages. The first stage is the exploration stage, which requires GPT-4 to explore different architectures in the search space as much as possible, avoid the output of the same architecture, and avoid falling into a local optimum. We transform the sequences output by GPT-4 into actual HGNNs and train them on the input dataset to evaluate their performance in a downstream task and add the generated architectures and their accuracy results into the prompt, asking GPT-4 to generate new architectures. 

After accumulating a part of the samples, we enter the second stage, i.e., the optimization stage. At this stage, we no longer need to explore the search space but require GPT-4 to analyze how to obtain better architectures based on the already evaluated architectures explored in the previous iterations. The specific optimization method is chosen by GPT-4, we only state in the prompt that it needs to be analyzed based on existing architectures before the output of new sequences of architectures.

\begin{algorithm}[t]
	\caption{GHGNAS search algorithm.}
	\label{alg:A}
	\begin{algorithmic}[1]
		\REQUIRE
        Heterogeneous graph $\mathcal{G}$;
		Search space $\mathcal{A}$;
        Evaluation metric  $c$;
        \# of iterations in exploration $T_e$ ; 
       \# of iterations in optimization $T_o$ ; 
		\ENSURE 
	    The best architecture $\alpha^*$ \\

		\STATE Model set $M\leftarrow []$, model  performance set $C \leftarrow [ ]$
        \STATE Generate prompt $P_s$, $P_d$based on $\mathcal{A}$ and $\mathcal{G}$
        \STATE $P_f = \emptyset$
    
        //Exploration Stage
        \FOR {i=1 to  $T_e$}  
        \STATE $M_i = \textbf{GPT-4}(P_t+P_d+P_s+P_e+P_f),C_i = c(M_i)$ 
        \STATE $M = M \cup M_i,C = C \cup C_i$
        \STATE Generate feedback prompt $P_f$ based on $M,C$ 
        \ENDFOR
        \STATE  $M',C'\leftarrow TopK(M,C)$ 
        \STATE Generate prompt $P_f$ based on $M'$ and $C'$
        
        //Optimization Stage
        \FOR{i=1 to  $T_o$}
            \STATE $M_i = \textbf{GPT-4}(P_t+P_d+P_s+P_o+P_f),C_i = c(M_i)$ 
        \STATE $M' = M' \cup M_i,C' = C' \cup C_i$
        \STATE Generate feedback $P_f$ prompt based on  $M',C'$
        \ENDFOR
         \STATE $M^*\leftarrow TopK(M \cup M')$ 
        \STATE Re-train $M^*$ and select the architecture with the best performance on the validation set as $\alpha^*$
		\RETURN $\alpha^*$
	\end{algorithmic}
\end{algorithm}

\subsection{Algorithms}
Algorithm 1 outlines the detailed process of GHGNAS. Initially, we construct the descriptions $P_s$ and $P_d$ of the search space and the input heterogeneous graph dataset. Lines 3 to 8 represent the exploration stage, where the algorithm guides GPT-4 to explore possible architectures in the search space by combining the prompts given in Table~\ref{prompt} and taking the existing architectures and their accuracy results as feedback $P_f$. At this point, the main purpose of adding feedback is to avoid generating repeated architectures. Line 9 shows the optimization stage, adds the current best-performing $n$ architectures to $P_f$, and modifies the strategy of this stage to $P_o$. We ask GPT-4 to analyze how to obtain better architectures based on these architectures and add newly generated architectures and their accuracy results to feedback $P_f$. At this point, the purpose of adding feedback is to verify whether the sampling preference for GPT-4 and the generated architectures are correct. If GPT-4 find that the sampling results are weaker than existing architectures, they will avoid using the same sampling preference and redo the analysis.

Due to the token limit of GPT-4, when it is unable to continue generating new architectures, we will select the currently top-k architectures as knowledge and add them to the prompt, and then continue to interact with GPT-4. In addition, the best result may not appear in the last iteration. Thus, after the search ends, we will read the interaction history to find the top architectures with the best performance, verifying their accuracy results. In the last step, we take the architecture with the highest accuracy result as the ultimate output.

\section{Experiments}
In the experiment, the large language model we use is GPT-4 version 2023-3-14. The experiments are designed to answer the following three questions:
\begin{itemize}
    \item Can GPT-4 be used for heterogeneous graph neural architecture search tasks?
    \item If GPT-4 works, compared with previous manually designed architectures and existing neural architecture search algorithms, how does the architecture designed by GHGNAS perform?
    \item How about the performance of the designed GHGNAS prompts, and how to optimize the prompts?
\end{itemize}

According to the above questions,  we set up three experiments:  
\begin{itemize}
\item First, we traverse a smaller search space as benchmark and compare the accuracy and ranking of GHGNAS with existing neural architecture search methods. 
\item Second, in a real-world architecture search environment, we compare GHGNAS with popular GNNs, HGNNs, and popular graph NAS methods. 
\item Third, we design ablation study and compare the performance of the variants of the proposed prompts on downstream graph learning tasks to prove the utility of our prompts.
\end{itemize}

\subsection{Experimental Setup}
\textbf{Datesets.} We evaluate our method on two tasks: node classification and recommendation. In the node classification task, we use two widely used academic network datasets, i.e., ACM and DBLP. The target node of the ACM dataset is Paper (P), and the target node of the DBLP dataset is Author (A). In the recommendation task, we use two datasets, i.e., Amazon and Yelp. The target edge of Amazon is User-Item, and the target edge of Yelp is User-Business.  The details of dataset division are the same as HGNAS.
\\
\textbf{Baselines.} First, we compare the most commonly used GCN~\cite{kipf2016semi} and GAT~\cite{velickovic2017graph} on homogeneous graphs, as well as the homogeneous graph architecture search method GraphNAS~\cite{gao2022graphnas++}. These methods require the conversion of heterogeneous graphs into homogeneous graphs as input. In addition, we also compare manually designed HGNNs, such as RGCN~\cite{schlichtkrull2018modeling}, HAN~\cite{wang2019heterogeneous}, HGT~\cite{hu2020heterogeneous}, and MAGNN~\cite{fu2020magnn}. We also consider two NAS methods, HGNAS~\cite{gao2021heterogeneous} and DiffMG~\cite{ding2021diffmg}. However, since the dataset and search space of DiffMG are different from ours, we supplemented the search space experiment of DiffMG in the appendix.
\\
\textbf{Hyper-parameters.}
For GHGNAS, the temperature of the GPT-4 is set to 1 during the exploration stage, and the temperature is set to 0 during the optimization stage. And we let GPT-4 generate 20 architectures in each iteration. Finally, we retrained the top 10 architectures based on their performance scores on the validation set. This retraining is conducted ten times to select the best-performing architecture.
We use HGNAS to sample 200 architectures for exploration and to sample 100 results for exploitation. For Random, we randomly sample 300 HGNNs from the search space. 
For HGNNs, the learning rate is set to 0.005, the dropout is set to 0.6, and the hidden dimension is set to 256.

\subsection{Experimental Results}
\begin{table}[b]
\caption{Experimental results (\%) on ACM and DBLP for node classification \textit{w.r.t.} macro-F1 on benchmark.}
\resizebox{\linewidth}{!}{
\begin{tabular}{c|cc|cc}
\hline
& \multicolumn{2}{c|}{ACM}& \multicolumn{2}{c}{DBLP}\\ 
& macro-F1& rank          & macro-F1& rank          \\ \hline
meta-path & 89.75& 1725          & 91.72
& 167           \\ 
All relations   & 91.18   & 245  & 90.38   & 240           \\ \hline\hline
Random         & 91.60     & 38           & 93.51 & 4             \\
HGNAS          & 91.52       & 64         & 93.14   & 25            \\
GHGNAS            & \textbf{91.70}   & \textbf{20}   & \textbf{93.52}  & \textbf{3}    \\ \hline\hline
Random (avg)   & 91.34  & 66.2    & 93.09      & 31.4 \\
HGNAS (avg)    & 91.27     & 256.4         & 92.41     & 101.2         \\
GHGNAS (avg)      & \textbf{91.58}   & \textbf{48.6} & \textbf{93.23}  & \textbf{20.4}       \\ \hline
\end{tabular}}
\label{bench}
\end{table}
\begin{table*}[h]
    \centering
    \caption{Experimental results (\%) on ACM and DBLP for node classification \textit{w.r.t.} macro-F1 and Experimental results (\%) on Amazon and Yelp for link prediction \textit{w.r.t.} AUC. `avg' represents the average performance of architectures obtained from five different searches. }
\resizebox{\linewidth}{!}{
\begin{tabular}{c|cc|cc|cc|cc}
\hline
 & \multicolumn{2}{c|}{ACM} & \multicolumn{2}{c|}{DBLP} & \multicolumn{2}{c|}{Amazon} & \multicolumn{2}{c}{Yelp} \\ \cline{2-9} 
 & Val & Test & Val & Test & Val & Test & Val & Test \\ \hline
GCN & 92.47±0.45 & 90.33±0.59 & 59.62±1.31 & 83.03±0.93 & 78.96±0.06 & 78.12±0.06 & 90.43±0.12 & 91.62±0.11 \\
GAT & 92.59±0.19 & 91.25±0.27 & 60.36±0.77 & 84.34±0.94 & 75.00±0.77 & 75.43±0.07 & 85.46±0.06 & 85.51±0.20 \\
RGCN & 92.12±0.49 & 91.25±0.50 & 86.77±0.59 & 93.66±0.76 & 74.77±0.10 & 73.85±0.12 & 88.93±0.19 & 88.63±0.17 \\
HAN & 93.20±0.42 & 91.16±0.27 & 85.01±0.29 & 92.12±0.20 & 74.37±0.37 & 75.96±0.23 & 89.35±0.23 & 89.64±0.20 \\
HGT & 89.12±0.66 & 88.93±0.38 & 84.92±0.43 & 92.97±0.20 & 75.69±0.76 & 74.30±0.02 & 91.22±0.15 & 91.73±0.20 \\
MAGNN & 92.13±0.39 & 91.30±0.38 & \textbf{87.13±0.32} & 93.57±0.20 & 75.24±0.38 & 75.90±0.33 & 91.34±0.20 & 90.50±0.30 \\ \hline
Random & 92.96±0.31 & 91.70±0.46 & 86.05±0.72 & 93.50±0.44 & 78.70±0.12 & 77.84±0.11 & 91.51±0.69 & 92.03±0.79 \\
Graphnas & 91.73±0.47 & 90.89±0.54 & 79.07±0.20 & 75.49±0.34 & 77.78±0.16 & 78.21±0.21 & 91.44±0.22 & 91.34±0.10 \\
HGNAS & 93.56±0.28 & 91.45±0.10 & 86.15±0.53 & 93.57±0.36 & 79.30±0.22 & 78.58±0.19 & 92.39±0.29 & 92.12±0.14 \\
GHGNAS & \textbf{93.69±0.50} & \textbf{92.37±0.19} & 86.72±0.47 & \textbf{93.84±0.21} & \textbf{79.90±0.18} & \textbf{79.16±0.15} & \textbf{92.78±0.16} & \textbf{92.80±0.15} \\ \hline\hline
Random (avg) & 92.29±0.56 & 90.89±0.74 & 85.46±1.20 & 93.24±2.13 & 78.28±1.29 & 77.52±1.37 & 90.57±2.25 & 91.03±1.74 \\
Graphnas (avg) & 91.27±1.09 & 90.71±1.02 & 76.63±1.71 & 73.30±1.82 & 77.72±0.35 & 77.64±0.32 & 90.70±0.44 & 91.22±0.29 \\
HGNAS (avg) & 92.57±0.34 & 91.21±0.63 & 85.53±1.69 & 91.35±1.36 & 79.19±0.40 & 78.50±0.39 & 92.20±0.35 & 91.96±0.21 \\
GHGNAS (avg) & \textbf{93.11±0.45} & \textbf{91.96±0.44} & \textbf{85.93±0.28} & \textbf{93.54±0.33} & \textbf{79.55±0.51} & \textbf{78.76±0.36} & \textbf{92.33±0.45} & \textbf{92.29±0.53} \\ \hline
\end{tabular}}
\label{real}
\end{table*}
\subsubsection{Results on the benchmark.}

We test GPT-4's generation capability on heterogeneous graph neural architecture search within a very small search space. In this search space, GNN operations only include $"zero"$ and $"gcn"$, and the aggregation function $HAggr$ only includes $"sum"$. Therefore, the goal of this experiment can be taken as finding the best meta-structures for a given heterogeneous graph. When the layer number of HGNNs is set to 2, the benchmark contains 8,192 possible HGNN architectures for ACM and   1,024 for DBLP. We traversed all possible HGNNs and calculated the average accuracy five times for each architecture. In this experiment, we ask GPT-4 to explore three iterations and then optimize two iterations, searching a total of 100 possible architectures. We use the Random method to search 100 HGNNs. The hyperparameters of HGNAS remain unchanged because it needs to train the controller. 

The results are shown in Table~\ref{bench}. Specifically, "All relations" represents the architecture using $"gcn"$ for all relations in Equation 4, and the "meta-path" represents the architecture uses $"gcn"$ only for the relations that constitute the meta-path while setting all other relations to $"zero"$. The table reports the performance of the best architecture designed based on the meta-path.

\begin{figure}[t]
  \centering
  \begin{subfigure}[b]{0.45\linewidth}
    \includegraphics[width=\textwidth]{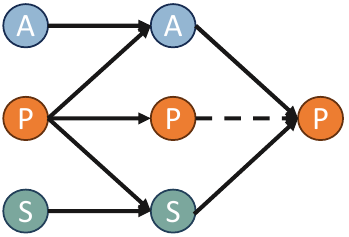}
    \caption{Designed HGNN on ACM. Authon(A), paper(P), subject(S)}
    \label{fig:sub1}
  \end{subfigure}
  \hfill
  \begin{subfigure}[b]{0.45\linewidth}
    \includegraphics[width=\textwidth]{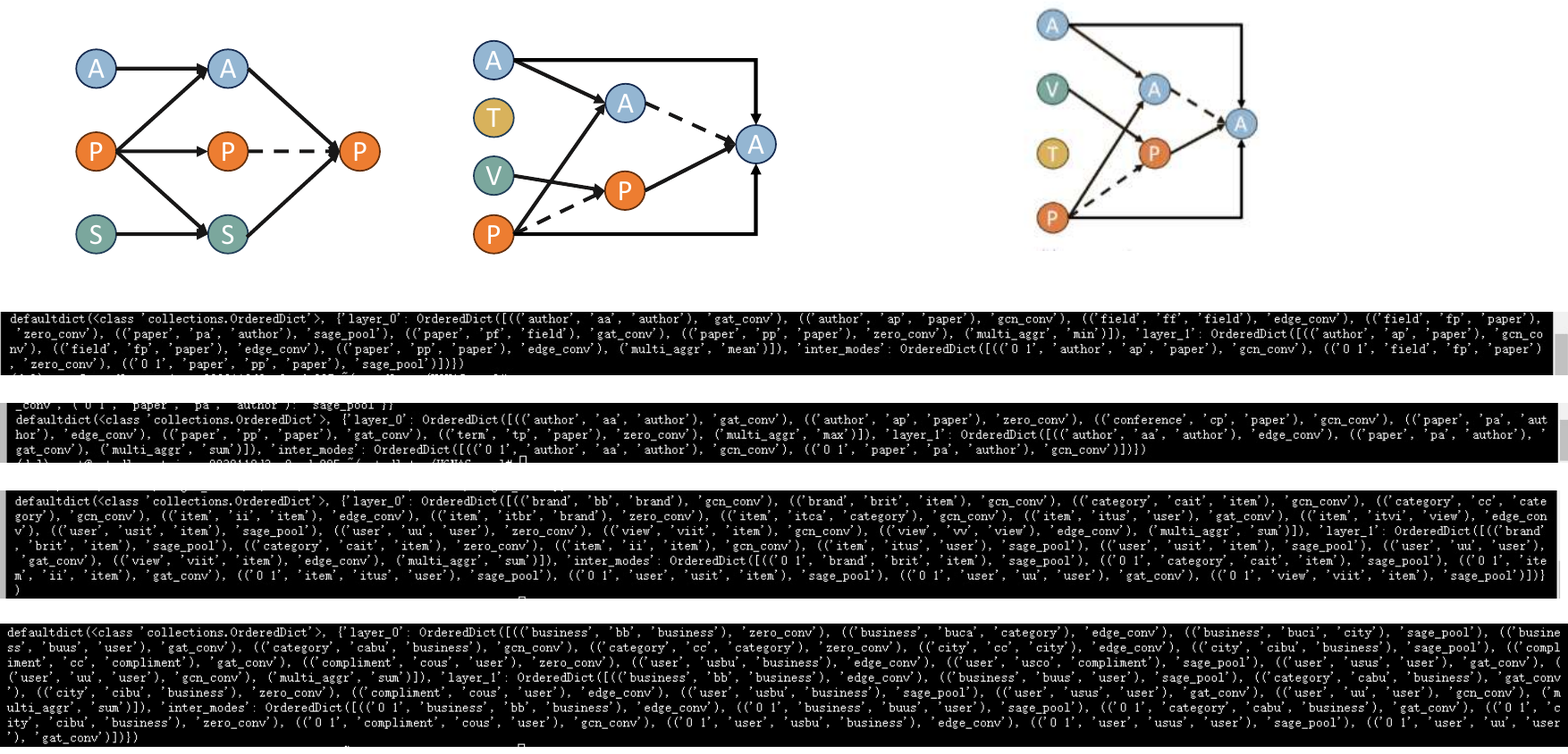}
    \caption{ Designed HGNN on DBLP. Authon(A), paper(P), term(T), venue(V)}
    \label{fig:sub2}
  \end{subfigure}
  \caption{HGNN designed by GHGNAS on the benchmark. 
  }
  \label{fig:hgnn on bench}
\end{figure}

Firstly, the architecture designed by the meta-path $PAP$ performs the best, but it still falls short of "All relations" on ACM. However, on DBLP, the performance of the architecture designed by the meta-path APV exceeds ‘All relations’ by 1.34\%.
Additionally, the NAS method can consider a more flexible message-passing method, which performs better than the architecture manually designed based on meta-path and the architecture using all relations.
Secondly, on both datasets, the architecture designed by GHGNAS outperforms other NAS methods. Furthermore, it exhibits the best average performance across five distinct searches. Notably, GHGNAS identified an architecture that ranks third on the DBLP dataset.

Figure \ref{fig:hgnn on bench} illustrates the HGNNs designed by GHGNAS. The left picture shows the best architecture found by GHGNAS on ACM, which includes the best meta-path PAP. Compared with the full connection, it discards A-P and S-P in the first layer. It also contains the best meta-path APV on DBLP and discards messages from T nodes.  This demonstrates that discarding part of messages within heterogeneous graphs can improve performances.

\subsubsection{Results on HGNAS search space.}
\begin{figure*}
  \centering
  \includegraphics[width=\textwidth]{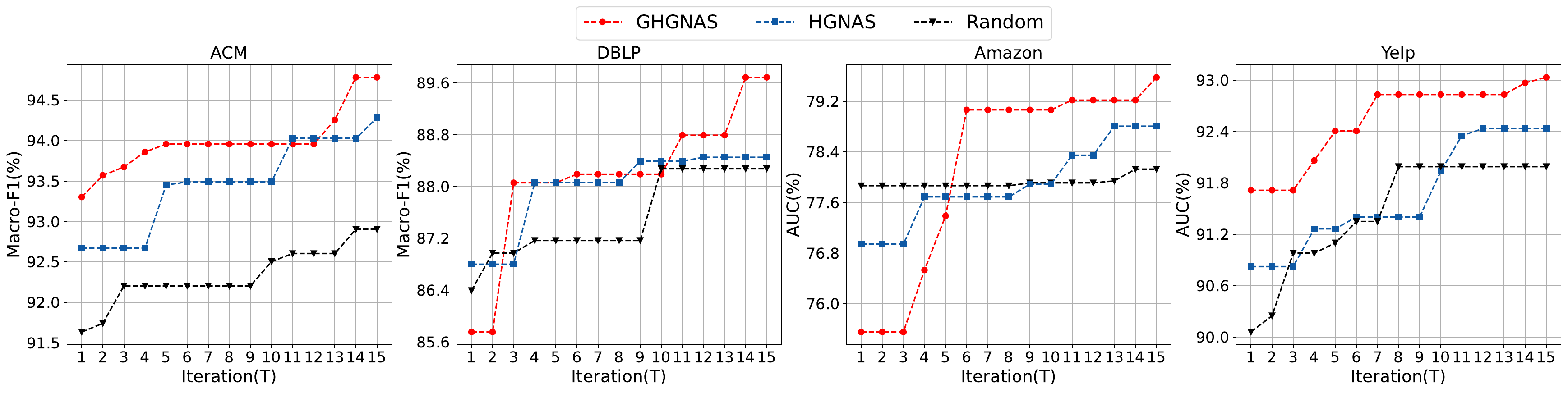}
  \caption{ The performance of the best HGNN model designed by GHGNAS and other baselines \textit{ w.r.t.} iterations. The HGNN model designed by GHGNAS outperforms others during the exploration stage. At the optimization stage, GHGNAS further improves the HGNN model.}
  \label{fig: best}
\end{figure*}
\begin{figure}
  \centering
  \includegraphics[width=\linewidth]{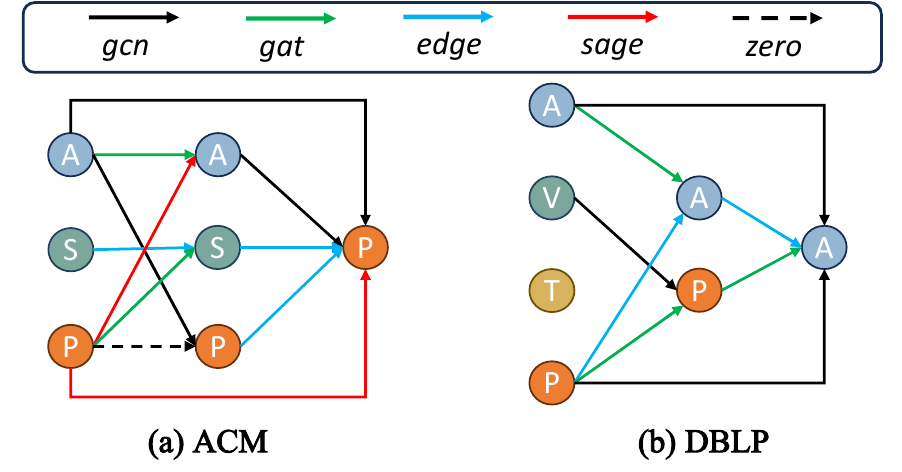}
  \caption{The best HGNN architecture designed by GHGNAS on the ACM (left) and DBLP (right). 
  }
  \label{fig:top_HGNN}
\end{figure}

According to the first part of the experiment, we confirmed that GPT-4 has the ability to complete the task of architecture search, so we considered continuing testing with complete HGNAS search space. In this part, we let GHGNAS iterate ten times using the exploration strategy, and then iterate five times using the optimization strategy. The left part of Table 4 shows the results of the node classification task on the ACM and DBLP datasets w.r.t. macro-F1 and the right part is the results of the recommendation task on the Amazon and Yelp datasets \textit{w.r.t.} AUC. Table \ref{real} provides the results of running five times on the test set.  In addition, Figure \ref{fig: best} shows the accuracy of the best architecture during iteration.

First, as shown in Table \ref{real}, manually designed HGNNs did not achieve good results, and GAT and GCN sometimes even surpassed some HGNNs. This shows that manually designed algorithms sometimes convey limited information or convey information that has a negative impact on the results.
Second, although the performance of HGNNs designed by GHGNAS is slightly weaker than MAGNN on the DBLP validation set, it is always better than other baselines on the test set. In addition, since the search process of GHGNAS is insensitive to seeds, it is always able to design excellent architectures in different search iterations, thus maintaining superior average performance. In Figure \ref{fig:top_HGNN}, the HGNN architecture newly designed for ACM still includes the optimal meta-path PAP, and the architecture for DBLP also includes the optimal meta-path APV. By employing a more extensive set of GNN operations, they also exhibit improved performance compared to the HGNNs designed in Figure \ref{fig:hgnn on bench}. This indicates that NAS, which selects GNNs operations for different relations, is advantageous for downstream tasks, and it also highlights that GHGNAS can design excellent architecture in complicated search spaces.
\subsubsection{Ablation study on different prompts.}
\begin{table}[b]
\caption{Results of different prompts.}
\resizebox{\linewidth}{!}{
\begin{tabular}{c|cc|cc}
\hline
 & ACM & DBLP & Amazon & Yelp \\ \cline{2-5} 
 & \multicolumn{2}{c|}{Macro-f1} & \multicolumn{2}{c}{AUC} \\ \hline
GHGNAS & 92.46±0.06 & 93.91±0.21 & 79.14±0.17 & 92.81±0.15 \\
no dataset & 92.21±0.36 & 93.76±0.33 & 78.03±0.13 & 92.66±0.08 \\
no operation & 90.75±0.40 & 93.31±0.64 & 72.38±0.14 & 91.72±0.25 \\
no strategy & 92.11±0.41 & 93.62±0.38 & 77.80±0.08 & 92.62±0.21 \\ \hline
\end{tabular}}
\label{abl}
\end{table}

\begin{figure*}
  \centering
    \includegraphics[width=\textwidth]{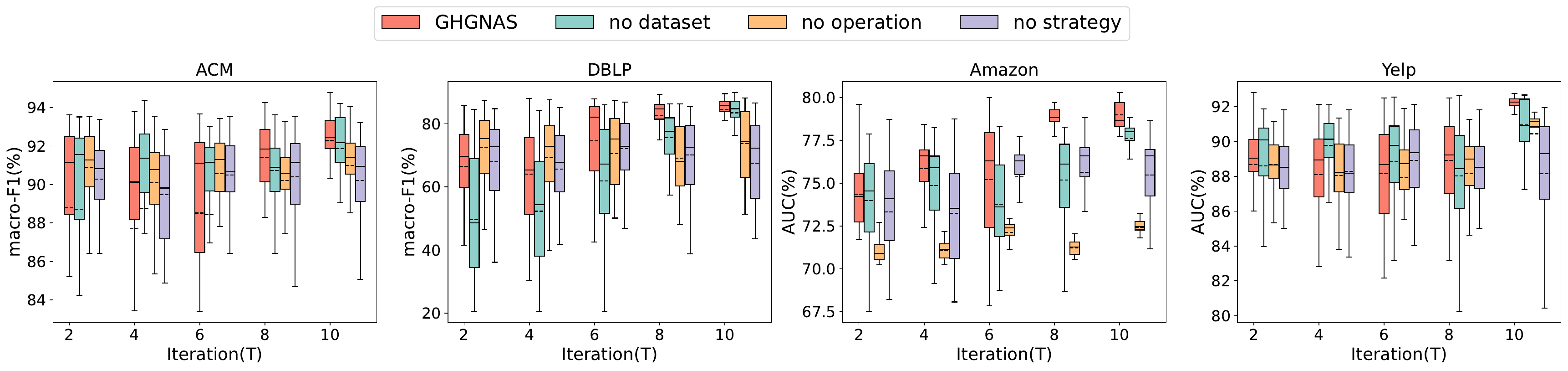}
  \caption{The performance of HGNN designed by different prompts \textit{w.r.t.} iterations.
  At the end of the search, GHGNAS  outperforms its variants.
  }
  \label{fig:avg}
\end{figure*}
In this part, we test the impact of different prompts. The prompts include task description, information about heterogeneous graphs, search space, search strategy, and feedback after verifying the architecture. We compare the impact of removing a part of the prompt on the results. The task description and feedback results need to be retained; otherwise, it will affect the generated content. When there is no information about the heterogeneous graph, after generating a sequence representing the architecture, replace the specific relationship with $edge_i$. When there is no search space operation description, 
we ask GPT-4 to choose a number for each element in a sequence of a specified length but do not tell what the numbers specifically mean. A total of five iterations of exploration and five iterations of optimization are conducted.

Table~\ref{abl} shows the best results obtained by different prompts. The results show that when a part of the description is removed, the outputs of GPT-4 suffer from an accuracy drop. Among them, removing operations from the search space has the most serious impact, with a decrease of 6.75\% on the Amazon dataset and a decrease of 1.71\% on the ACM dataset. Figure~\ref{fig:avg} shows the mean accuracy and variance of the designed HGNNs during the iteration process. Without the description of the search strategy, even if a batch of better architectures has been obtained in previous iterations, it is impossible to use these architectures for prompt optimization. Thus, the average performance always fluctuates. 
When there is no description of the operations in the search space, the average performance on Amazon lags behind other variants. In addition, even in the optimization stage, the improvement is insignificant on ACM and DBLP.
When there is no information about the input heterogeneous graphs, although the average performance is not as good as the original prompt, GHGNAS can still optimize architectures effectively, with the average results of the generated architectures improving steadily. These observations show that GPT-4 has good generalization capability and is capable of running neural architecture search tasks effectively even if it does not know the specific information of input datasets. 

\subsection{Discussions}
In this section, we discuss based on the above experimental results the advantages and limitations of the models, the prompts, and interpretability. 

Firstly, our method is capable of solving the heterogeneous graph neural architecture search task and performs better than existing HGNNs and HGNAS models. The experimental results show the HGNNs designed by GHGNAS are stably better than HGNAS based on reinforcement learning, which shows that our method has a better ability to perform neural architecture search. Compared with DiffMG, which is largely affected by the selected seeds, GHGNAS can search for accurate architectures in different conversations, and a batch of results with excellent performance can be obtained during the iterative process. This shows that using GPT-4 to complete architecture search tasks greatly reduces the requirements for professional knowledge and solves the pain points of previous neural architecture search methods. As users, we only need to accurately describe the task.

Secondly, the results generated by GHGNAS are interpretable. In our task, GPT-4 is used as a black box optimizer to complete the search and optimization of the architecture and has achieved better performance than the traditional interpretable neural architecture search solution. This results in that we only know that the architecture of GHGNAS search is available, but cannot explain the decision-making process. However, we can ask GPT-4 to give its own analysis results and sampling preference in the prompt. An example of a response from GPT-4 is as follows: 
\textit{\begin{itemize}
    \item  The operation $"gcn"$ frequently appears in high-performance architectures. Thus,  this operation is likely to have a strong correlation with model performance, particularly when the edge is (A-P), (S-P), and (P-P).
    \item The operation $"gat"$  also performs very well, appearing many times in the best architectures, which implies that $"gat"$  works very well for diverse layers.
    \item The operation $"sage"$ rarely appears, particularly at the last entry of the action list, which indicates that the operator may not perform as expected in the benchmark datasets.
    \item The $"edge"$ and $"zero"$, while appearing less frequently, are often found in high-performing actions as well, indicating that they have a positive impact on the architecture in specific combinations.
    \item The models with accuracy above 0.923 use a diversified search strategy, showing that different types of convolutions are beneficial for the classifier.
    \item The operation $"gcn"$ and various convolutions can improve model performance. It is suggested to avoid using $"sage"$ frequently. However, it might have a potential advantage when combining $"sage"$  with other operations.
\end{itemize}}

\begin{figure}[t]
    \centering
    \includegraphics[width=\linewidth]{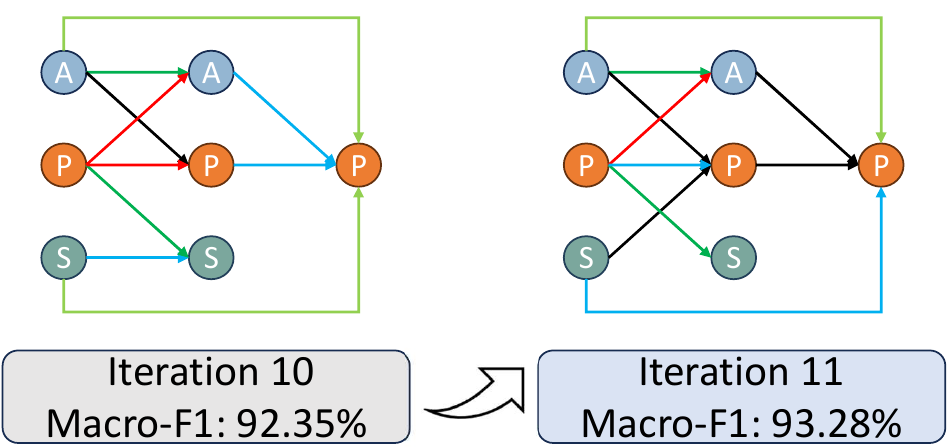}
    \caption{An example of the GHGNAS optimization stage on ACM. 
    The left part shows the HGNN designed by GHGNAS after ten iterations using the exploration strategy. After entering the optimization stage, in the 11th iteration, the newly generated HGNN achieved an improvement of 0.93\%.
    }
    \label{fig:6}
\end{figure}

Figure \ref{fig:6} shows an optimization example of GHGNAS. The left part shows an architecture designed in the exploration stage. GPT-4 based on the analysis results samples a new architecture that uses more $gcn$ operations and achieves a 0.93\%  improvement on the validation set. This means that GPT-4 has the ability to optimize the architecture.

Third, whether GPT-4 already knows the best architecture for the data set. This is a concern raised by GENIUS~\cite{zheng2023can}. If the pre-training process of GPT has already included the optimal architecture of a specific dataset, then using GPT for neural architecture search is a process of directly finding answers based on pre-training knowledge. The experiments in this paper can address this concern. First, HGNAS~\cite{gao2021heterogeneous} was introduced in 2021, at a time when GPT did not possess knowledge of HGNAS. Secondly, when we do not provide information on heterogeneous graphs, GPT-4 can still design excellent architectures for the four data sets. It can be seen that GPT-4 observes and optimizes the performance of different architectures, rather than directly generating excellent architectures based on existing knowledge in pre-training.

\vspace{3pt}

\textbf{\large{Limitations}.} First, one of the pain points of using GPT-4 for heterogeneous graph neural architecture search is that the usage of OpenAI’s API  is charged. The understanding and generalization capabilities of GPT-3.5 or other LLMs are still far behind GPT-4, they prefer to provide the program code that generates the neural architecture rather than providing the architecture itself. In the future, we will add other powerful yet free LLMs to the tasks.

Second, the reproducibility of the optimization process is limited. Although the process has interpretability, the optimization is not the same. Even with the same prompt, there is some randomness in the content generated by GPT-4 each time. This is also one of the reasons why in the later stages, it is possible to generate an architecture with poor performance.

Third,  GHGNAS still relies on a manually designed search space. We hope that in the future, GHGNAS can be used to complete the tasks of search space formulation and exploration at the same time, to the greatest extent reducing the dependence of heterogeneous graph neural architecture search on knowledge and experience. 

\section{CONCLUSIONS}
In this paper, we introduced a new Heterogeneous Graph Neural Architecture Search model GHGNAS, which uses the large language model GPT-4 to complete the task of heterogeneous graph neural architecture search. Experimental results show that GHGNAS can use the generation and decision-making capabilities of GPT-4 to find new HGNN architectures that are better than existing models based on reinforcement learning and differentiable search algorithms. Just by modifying the prompts, GHGNAS can be migrated to new search spaces and perform well.
 GHGNAS alleviates the dependence on professional knowledge and the unstable search results of existing heterogeneous graph search algorithms. 
In future work, we hope to use the capabilities of large language models to solve a more challenging problem of generating new search spaces for heterogeneous graph neural architecture search.
\bibliographystyle{ACM-Reference-Format}
\bibliography{ref}

\appendix

\section{Appendix}
\subsection{Comparison with DiffMG}
DiffMG~\cite{ding2021diffmg} employs a differentiable search algorithm with the aim of automatically searching for the meta-structures that are applicable to heterogeneous graphs and downstream tasks, thereby capturing semantic information that is more complex than meta-paths. In experiment 5.2.1, we only considered the meta-structures found by DiffMG and did not evaluate the performance of GHGNAS in the DiffMG search space. We redesigned the prompts based on DiffMG’s tasks and search space, and further compare them in the tasks of ACM and DBLP node classification. 

\subsubsection{Search space}
DiffMG defines the search space as a Directed Acyclic Graph (DAG), where nodes on the DAG represent the intermediate states of the message-passing process, and edges represent candidate edge types. Furthermore, DiffMG constrains the search space based on the target node, which can be specifically expressed in the following form:
\begin{equation}
\mathcal{A}_{j,i}=\begin{cases}\mathcal{A}\cup\{I\}&j<N\text{and}i=j-1\\\mathcal{A}\cup\{I\}\cup\{O\}&j<N\text{and}i<j-1\\\overline{\mathcal{A}}&j=N\text{and}i=N-1\\\overline{\mathcal{A}}\cup\{I\}\cup\{O\}&j=N\text{and}i<K-1\end{cases}.
\end{equation}
where $\mathcal{A}$ represents the adjacency matrix of the candidate edge type, $I$ and $O$ represent the identity matrix and the all-zero matrix respectively. When it reaches the last node of the DAG, it can only choose the candidate edges that are currently related to the task. This method reduces the search space and filters out candidate edge types that have no impact on the final node representation, thereby improving search efficiency. For more specific information, you can refer to the original paper.
\subsubsection{Prompts for the DiffMG search space}
Taking the ACM data set as an example, we modify the prompt and use natural language to describe the search space and constraints. Similarly, GHGNAS can also guide GPT-4 in the prompt to choose candidate edges that are relevant to the task target. Specifically, the design of the prompt is also similar to that in the HGNAS search space, we use sequences to represent meta-structures and require GPT-4 to select a candidate edge type for each element on the sequence. The specific details are as follows: 

\begin{tcolorbox}
Task description $P_t$:\\
Our task is to find a meta-structure of heterogeneous graphs that maximizes accuracy on downstream tasks.\\
Search space $P_s$:\\
For a given meta-structure \{H0-H1, H1-H2, H2-H3, H3-H4\} \{H0-H2, H0-H3, H1-H3, H0-H4, H1-H4, H2-H4\}, H0-H1, H1-H2, H2-H3 should be selected from [PA, AP, PS, SP, I], H3-H4 should be selected from [PA, PS, I] and H0-H4, H1-H4, H2-H4 should be selected from [PA, PS, I, O] because the target is P. For H0-H2, H0-H3, and H1-H3 you can choose from [PA, AP, PS, SP, I, O]\\
\end{tcolorbox}
The rest of the part is consistent with the template given in the text. By simply modifying the template, GHGNAS can be migrated to heterogeneous graph meta-structure search tasks without the need to design additional search algorithms.
\begin{table}[t]
\small
\caption{Results of node classification in DiffMG' search space w.r.t. macro-F1.}
\resizebox{\linewidth}{!}{
\begin{tabular}{cc|cc|cc}
\hline
 &  & \multicolumn{2}{c|}{DBLP} & \multicolumn{2}{c}{ACM} \\ \hline
\multicolumn{1}{c|}{\multirow{6}{*}{DiffMG}} & seed & Val & Test & Val & Test \\ \cline{2-6} 
\multicolumn{1}{c|}{} & 0 & 95.30±0.29 & \textbf{94.16±0.10} & 92.26±0.43 & 91.30±0.27 \\
\multicolumn{1}{c|}{} & 1 & 94.70±0.73 & 92.58±0.20 & 91.90±0.45 & 91.42±0.15 \\
\multicolumn{1}{c|}{} & 2 & 82.63±0.10 & 81.48±0.30 & 93.34±0.63 &  92.07±0.10\\
\multicolumn{1}{c|}{} & 3 & 81.00±0.67 & 82.95±0.44 & 90.51±0.85 & 89.55±0.11 \\
\multicolumn{1}{c|}{} & 4 & 83.37±0.48 & 81.93±0.21 & 92.63±0.47 & 91.30±0.21 \\ \hline
\multicolumn{2}{c|}{GHGNAS} & \textbf{96.00±0.36} & 93.92±0.16 & \textbf{93.55±0.58} & \textbf{92.21±0.21} \\ \hline\hline
\multicolumn{2}{c|}{DiffMG (avg)} & 87.40±6.26 & 86.62±5.56 & 92.13±0.94 & 91.13±0.93 \\ 
\multicolumn{2}{c|}{GHGNAS (avg)} & \textbf{95.42±0.32} & \textbf{93.37±0.25} & \textbf{93.05±0.38} & \textbf{91.80±0.52} \\ \hline
\end{tabular}}
\label{vsdiff}
\end{table}
\subsubsection{Experimental results.}
We follow all the default settings of DiffMG, and GHGNAS uses the exploration strategy and optimization strategy for five iterations each. The results indicate that GHGNAS surpasses DiffMG in terms of node classification task, with the exception of a 0.24\% lag on the DBLP test set. Moreover, in line with the experimental findings in the main text, DiffMG can only yield satisfactory results when utilizing specific seeds, whereas GHGNAS is capable of discovering superior architectures following ten iterations across various search processes.

\subsection{STATISTICS OF DATASETS}
The experiments in this paper use the same datasets and the partitioning method as in HGNAS~\cite{gao2021heterogeneous}, and the dataset information is shown in Table~\ref{dataset}. For the ACM and DBLP datasets, we use 400 nodes for training and 400 nodes for validation. For Yelp and Amazon, we consider ratings higher than three as positive pairs, and the rest as negative pairs. Then, we randomly divide 50\% of the positive pairs into training, validation, and test sets at a ratio of 3:1:1. For the comparison with DiffMG~\cite{ding2021diffmg}, the experimental settings used can be referred to in the original paper.
\begin{table}[t]
\caption{Dataset statistics.}
\resizebox{\linewidth}{!}{
\begin{tabular}{c|c|ccc}
\hline
Dataset & Relations(A-B)    & \#A   & \#B   & \#A-B  \\ \hline
ACM     & Author-Paper(A-P)      & 17,351 & 4,025  & 13,407  \\
        & Paper-Subject(P-S)     & 4,025  & 72    & 4,025   \\ \hline
DBLP    & Author-Paper(A-P)      & 4,057  & 14,325 & 19,645  \\
        & Paper-Term(P-T)        & 14,325 & 7,723  & 85,810  \\
        & Paper-Venue(P-V)       & 14,325 & 20    & 14,328  \\ \hline
Amazon  & User-item(U-I)         & 6,170  & 2,753  & 195,791 \\
        & Item-View(I-V)         & 2,753  & 3,857  & 5,694   \\
        & Item-Category(I-C)     & 2,753  & 22    & 5,508   \\
        & Item-Brand(I-B)        & 2,753  & 334   & 2,753   \\ \hline
Yelp    & User-Business(U-B)     & 16,239 & 14,284 & 198,397  \\
        & Business-City(B-Ci)     & 14,285 & 47      &  14,267      \\
        & User-User(U-U)         & 16,239      &  13269     & 158,590       \\
        & User-Compliment(U-Co)   & 16,239      &  11     &   76,875     \\
        & Business-Category(B-Ca) & 14,284      &  511     &  40,009      \\ \hline
\end{tabular}}
\label{dataset}
\end{table}

\begin{figure}[h]
  \centering
  \begin{subfigure}[h]{0.45\linewidth}
    \includegraphics[width=\textwidth]{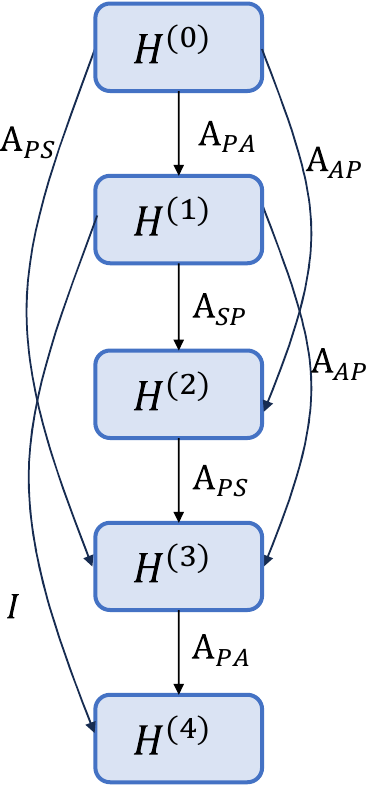}
    \caption{ACM}
    \label{fig:diffmg_acm}
  \end{subfigure}
  \begin{subfigure}[h]{0.45\linewidth}
    \includegraphics[width=\textwidth]{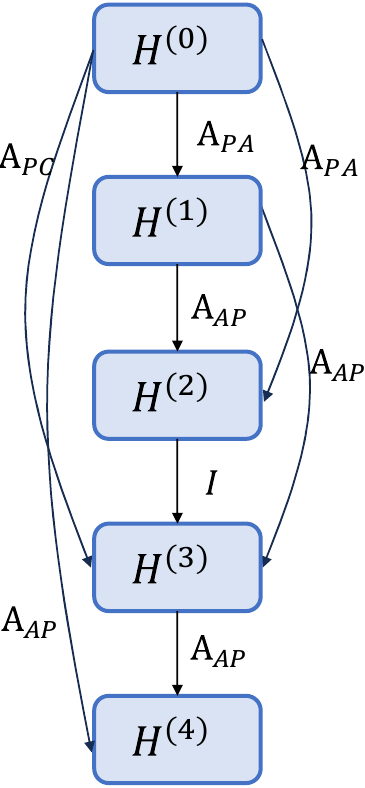}
    \caption{DBLP}
    \label{fig:diffmg_dblp}
  \end{subfigure}
  \caption{Meta-structures designed by GHGNAS.}
  \label{fig:DiffMG}
\end{figure}

\end{document}